\documentclass[10pt]{article}
\usepackage[preprint]{tmlr}
\usepackage{microtype}
\usepackage{amsmath,amssymb,amsthm,mathtools}
\usepackage{booktabs,array,tabularx,url,graphicx}
\usepackage{algorithm}
\usepackage{algpseudocode}
\usepackage{tikz}
\usetikzlibrary{arrows.meta,positioning}
\usepackage[hidelinks]{hyperref}
\hypersetup{pdftitle={Evidence-Governed Adoption of Self-Generated Updates in Autonomous AI Agents},pdfauthor={Subodh Lonkar},pdfsubject={Preprint}}

\newtheorem{definition}{Definition}
\newtheorem{assumption}{Assumption}
\newtheorem{theorem}{Theorem}
\newtheorem{proposition}{Proposition}
\newtheorem{corollary}{Corollary}
\newtheorem{remark}{Remark}

\newcommand{\E}{\mathbb{E}}
\newcommand{\Prob}{\mathbb{P}}

\newcommand{\Accept}{\operatorname{Accept}}

\title{Evidence-Governed Adoption of Self-Generated Updates in Autonomous AI Agents}
\author{\name Subodh Lonkar\\\addr Independent Researcher}
\def\openreview{\url{https://openreview.net/}}
\begin{document}
\maketitle
% TMLR requires disclosure of LLM assistance. Keep the title unmarked and
% place the disclosure as an unnumbered first-page footnote instead.
\begingroup
\renewcommand{\thefootnote}{}
\footnotetext{The author used ChatGPT as an assistive tool for research planning, literature discovery, conceptual organization, drafting and revision of explanatory text, and language refinement. The author is responsible for verifying the manuscript's sources, mathematical arguments, experimental code and results, and all claims and conclusions.}
\endgroup

\begin{abstract}
Autonomous agents increasingly modify prompts, memories, tool configurations, workflows, and control policies in response to interaction histories. The ability to generate a plausible modification, however, does not establish that it should become persistent system state. Candidate selection is adaptive, evaluators may be imperfect or manipulable, and repeated local improvements can conceal degradation relative to an earlier operating baseline. This paper develops Evidence-Governed Evolution (EGE), a statistical and operational framework for governing persistent adoption of self-generated agent modifications. EGE separates candidate generation from confirmatory adoption, evaluates frozen candidates on fresh paired audit samples, requires practically meaningful improvement on a prespecified primary objective, protects declared capabilities through reference-relative non-inferiority tests, and applies mandatory governance constraints before bounded canary deployment. The statistical contribution is an evaluator-calibrated simultaneous lower-bound construction for a countable sequence of adaptively generated candidates. It combines candidate-wise Hoeffding bounds with independent, candidate-specific calibration samples that compare the operational evaluator with a prespecified reference evaluation procedure. A second result formalises reference replacement through explicit bridge tests and derives a cumulative degradation bound along a versioned reference lineage. We also give a conservative audit-planning condition that links effect-size margin, evaluation sample size, calibration sample size, and a lower bound on acceptance probability. A reproducible Monte Carlo stress test illustrates the importance of calibration under candidate-dependent evaluator optimism; it is explicitly a stylised statistical experiment, not an evaluation of a deployed agent. The paper does not claim empirical validation on a real autonomous-agent system. All guarantees are conditional on bounded paired scores, fresh sampling, a valid reference-evaluation construct, and the declared evaluation distributions; they do not establish universal agent safety or robustness to arbitrary distribution shift.
\end{abstract}

\noindent\textbf{Keywords:} autonomous agents; self-evolving agents; adaptive data analysis; sequential inference; evaluator calibration; non-inferiority; AI governance.

\section{Introduction}
Autonomous agents are systems whose behaviour depends not only on foundation-model weights but also on prompts, memory, tools, workflows, planning procedures, routing policies, and execution constraints. An agent can therefore change its behaviour without modifying its underlying model. Candidate changes may be generated from execution traces, user feedback, self-reflection, search, or optimisation.

A candidate that performs well on recently encountered tasks is not necessarily a reliable improvement. It may overfit to development data, reduce performance on previously supported tasks, increase latency or resource use, or violate operational constraints. The risk is amplified when an agent repeatedly proposes and deploys changes with limited human intervention.

Prior work studies reflection, iterative refinement, and skill acquisition in language-model agents \citep{shinn2023reflexion,madaan2023selfrefine,wang2023voyager}. Automated prompt optimisation and evolutionary search provide additional candidate-generation mechanisms \citep{guo2024evoprompt}. These methods motivate the ability to generate changes; they do not by themselves settle whether a proposed change should become persistent system state. Broader work on AI safety identifies specification failures, distributional failures, and unsafe exploration as recurring challenges \citep{amodei2016concrete,garcia2015safe}.

The adoption problem has several properties that make it different from ordinary offline model selection. First, candidate proposals are often adaptive: the next proposal may depend on previous successes, failures, evaluator feedback, or deployment traces. Second, the candidate can modify the mechanism that will later produce evidence, for example by changing a prompt, retrieval policy, tool-selection rule, or evaluator-facing output format. Third, the system may optimise a narrow objective while being expected to preserve a much broader capability profile. Fourth, deployment is not a purely statistical event. A candidate can pass a benchmark and still fail under production load, interact unexpectedly with tools, or create external side effects that cannot be undone by restoring an earlier configuration.

These properties motivate a layered adoption design. The proposal mechanism is permitted to explore the design space, but it does not control the confirmatory audit process. The audit process measures a frozen candidate against prespecified comparators and margins. A governance layer enforces requirements that must not be traded against average task performance. Finally, an operational deployment layer limits exposure while collecting evidence about behaviour in the target environment. The layers are related but not interchangeable: a confidence bound cannot replace a permission boundary, and a canary cannot repair an invalid statistical comparison.

The scope of EGE is deliberately broader than parameter-level continual learning. A candidate may change a prompt, memory schema, retrieval strategy, tool interface, workflow, routing rule, or orchestration policy. The same principles can be applied to model-weight changes, but those changes may require additional controls for training-data provenance, model integrity, supply-chain security, and compatibility. EGE does not specify a universal metric set; metric selection remains an application-specific modelling and governance decision.

This paper addresses the adoption decision through Evidence-Governed Evolution (EGE). EGE separates candidate generation from evidence-based adoption and combines: (i) fresh audit data collected after a candidate is frozen; (ii) meaningful improvement on at least one prespecified primary objective; (iii) non-inferiority on every prespecified protected capability relative to a stable reference; (iv) mandatory governance checks; and (v) bounded canary deployment before persistent commitment.

The paper makes four bounded contributions. First, it formalises persistent adoption as a distinct decision problem for adaptively generated agent configurations, with separate development, calibration, confirmatory-audit, and deployment stages. Second, it gives a simultaneous lower-confidence-bound construction that accounts for candidate-dependent evaluator discrepancy using independent calibration data rather than assuming a known calibration allowance. Third, it formalises reference replacement through bridge comparisons and derives a cumulative capability-degradation bound along the reference lineage; this result makes explicit the statistical price of changing the protected baseline. Fourth, it provides an audit-planning inequality and a reproducible Monte Carlo stress test that exposes a failure mode of evaluator-naive acceptance. The Monte Carlo experiment is not presented as evidence about a real agent system. A real-agent evaluation remains necessary to establish operational usefulness and is identified as an outstanding empirical requirement.

The distinction from prior work is deliberately narrow. EGE does not introduce a new concentration inequality, a new general-purpose sequential-testing principle, or a complete agent safety architecture. Its contribution is a specific composition and formalisation for persistent agent-update adoption: candidate-wise confirmatory testing, candidate-specific evaluator calibration, protected-capability reference lineage, and an operational commitment protocol. The novelty claim is therefore about this problem formulation and the resulting guarantees, not about the underlying Hoeffding or union-bound tools.

The results are conditional statistical guarantees, not a proof of universal agent safety. In particular, rollback restores configuration but cannot reverse irreversible external actions.

\section{Related Work}
\subsection{Reflection, Refinement, and Persistent Skills}
Reflexion uses verbal feedback and episodic memory to improve subsequent agent attempts \citep{shinn2023reflexion}. Self-Refine iteratively critiques and revises model outputs \citep{madaan2023selfrefine}. Voyager accumulates reusable skills through interaction with an environment \citep{wang2023voyager}. EGE is complementary: these methods can generate candidate changes, whereas EGE governs whether a candidate is retained persistently.

EvoPrompt connects language models with evolutionary algorithms for discrete prompt optimisation \citep{guo2024evoprompt}. Such optimisation makes the distinction between development evidence and final audit evidence especially important: the data used to select a candidate should not also be treated as an untouched confirmation sample.

Recent surveys organise self-evolving agents by the components and stages of evolution, examine self-improvement in modern agentic systems, and study endogenous observations generated within agent interaction loops \citep{gao2026survey,ren2026survey,zhang2026endogenous}. These works motivate reliability concerns around self-generated learning signals; EGE focuses on the statistical and operational gate at adoption time.

\subsection{Adaptive Data Analysis and Sequential Inference}
Hoeffding's inequality gives finite-sample concentration bounds for independent bounded observations \citep{hoeffding1963}. Repeated adaptive reuse of a holdout can invalidate ordinary fixed-sample inference; reusable-holdout methods address this problem through controlled information release \citep{dwork2015reusable}. Confidence sequences provide time-uniform inference when data are monitored repeatedly \citep{howard2021confidence}. Classical group-sequential designs and alpha-spending methods likewise address repeated interim testing \citep{lan1983,demets1994}. Conformal risk control offers a complementary approach to controlling expected monotone losses under its own exchangeability and procedure-specific assumptions \citep{angelopoulos2024conformal}.

EGE's simple construction is a union-bound allocation across candidate indices and metrics. It is valid when each candidate is frozen before a fresh audit sample is collected and the stated conditional sampling assumptions hold. It does not permit unrestricted reuse or repeated optional inspection of a fixed sample.

\subsection{Safe Adaptation and Risk Constraints}
Safe policy improvement studies evidence-supported policy changes relative to a baseline \citep{laroche2019spibb}. Broader safe reinforcement-learning research considers performance and safety constraints during learning and deployment \citep{garcia2015safe,chow2018risk}. EGE extends the governance perspective to heterogeneous agent configuration changes and multiple protected capabilities. Its non-inferiority tests do not imply that every possible failure mode is controlled.

\subsection{Agent Self-Evolution Evaluation and the Adoption Gap}
Recent work has begun to study self-evolving agents as a distinct research area. Gao et al. organise the field around what, when, and how an agent evolves, while Ren et al. survey self-improvement mechanisms in modern agentic systems \citep{gao2026survey,ren2026survey}. Zhang et al. analyse endogenous observations as learning signals generated by an agent's own generation, evaluation, and interaction processes \citep{zhang2026endogenous}. Benchmarking work such as EvoAgentBench evaluates ability transfer during agent self-evolution \citep{gao2026evoagentbench}. Lin et al. distinguish the ability to produce useful harness updates from the ability of a task-solving agent to benefit from those updates, underscoring that update generation and update utility are separate capabilities \citep{lin2026harness}. These studies strengthen the motivation for evaluating agent evolution directly. EGE addresses a different stage: after a candidate update has been generated, what evidence should be required before it is allowed to alter persistent system state?

\begin{table}[t]
\centering
\small
\caption{Conceptual positioning of EGE relative to representative research streams. Entries describe primary emphasis, not an exhaustive claim that a method cannot be combined with another control.}
\label{tab:positioning}
\begin{tabularx}{\linewidth}{p{0.22\linewidth} X X}
\toprule
Research stream & Primary question & Relationship to EGE \\
\midrule
Reflection and iterative refinement & How can an agent improve a response or future attempt? & Can supply candidate modifications; does not by itself define persistent-adoption evidence. \\
Prompt and evolutionary optimisation & How can candidate prompts or configurations be searched? & Supplies candidate generation; development scores must not be reused as untouched confirmation evidence. \\
Safe policy improvement & When is a policy change supported relative to a baseline? & Provides related baseline-constrained reasoning; EGE targets heterogeneous agent-configuration updates and multiple protected capabilities. \\
Adaptive data analysis and confidence sequences & How can inference remain valid under repeated or adaptive analysis? & Supplies statistical tools; EGE instantiates a candidate-wise fresh-audit design and explicit calibration allocation. \\
Agent self-evolution benchmarks & Do self-generated experiences or harness updates improve agent capability? & Supplies evaluation settings; EGE governs whether a particular update is persistently adopted. \\
Canary deployment and release governance & How can a change be exposed gradually and monitored? & Supplies operational containment; EGE separates this from offline statistical evidence. \\
\bottomrule
\end{tabularx}
\end{table}

The table is a conceptual map, not a systematic review or a claim of exhaustive coverage. In particular, safe policy improvement, adaptive inference, and deployment canary techniques are mature research areas. EGE must therefore be judged by whether the combination yields useful, correctly stated guarantees for this adoption problem, rather than by treating the combination of familiar components as novelty in itself.

\section{Problem Formulation}
\subsection{Agent Configuration and Candidate Generation}
Let $I_j$ denote the incumbent configuration immediately before candidate index $j$ is assessed. Its components are represented by
\begin{equation}
I_j=(M_j,P_j,\mathcal{M}_j,T_j,C_j,G_j),
\label{eq:agent}
\end{equation}
where $M_j$ is the model, $P_j$ the prompt and policy configuration, $\mathcal{M}_j$ persistent memory, $T_j$ tools, $C_j$ orchestration logic, and $G_j$ governance constraints. A subset of these components may be immutable.

Given experience $\mathcal{E}_j$, a candidate generator produces
\begin{equation}
\widetilde A_j=\mathcal U(I_j,\mathcal E_j).
\label{eq:candidate}
\end{equation}
The generator may be an optimiser, an LLM-based editor, a search procedure, or a human-supervised process. EGE does not prescribe the generator.

\subsection{Metrics and Paired Estimands}
Let $\mathcal J$ be a nonempty finite set of primary objectives and $\mathcal K$ a finite set of protected capabilities. Their roles, score definitions, and evaluation distributions must be declared before audit outcomes are observed. Scores are normalised to $[0,1]$, so paired score differences lie in $[-1,1]$. The reference-evaluation procedure defines the target estimand used in the statistical guarantees; it need not be interpreted as error-free ground truth.

For metric $r$, let $q^R_r(A,Z)$ denote the reference-procedure score of configuration $A$ on instance $Z$. Define
\begin{equation}
D^R_{j,r}(Z)=q^R_r(\widetilde A_j,Z)-q^R_r(B_{j,r},Z),
\label{eq:paired}
\end{equation}
where
\[
B_{j,r}=\begin{cases}
I_j,&r\in\mathcal J,\\
A_{\mathrm{ref}},&r\in\mathcal K.
\end{cases}
\]
Thus primary objectives are compared with the incumbent, while protected capabilities are compared with a fixed, versioned reference. If a metric belongs to both sets with different comparators or thresholds, the two roles are represented as distinct tests.

The target difference is
\begin{equation}
\Delta^R_{j,r}=\E_{Z\sim\mathcal D_r}
[D^R_{j,r}(Z)\mid\mathcal H_j],
\label{eq:delta}
\end{equation}
where $\mathcal H_j$ is the information available immediately before candidate $j$ is frozen and its fresh samples are drawn. The distribution $\mathcal D_r$ is the prespecified target distribution for metric $r$. For a single configuration, define $Q^R_r(A)=\E_{Z\sim\mathcal D_r}[q^R_r(A,Z)]$; paired differences are therefore differences of these reference-procedure expectations when the same distribution is used.

\subsection{Capability Representation}
For exposition, an application may organise its evaluation metrics into a capability vector such as
\begin{equation}
\mathbf Q(A;\mathcal D)=
\begin{bmatrix}
Q_{\mathrm{task}} & Q_{\mathrm{gen}} & Q_{\mathrm{ret}} &
Q_{\mathrm{rel}} & Q_{\mathrm{eff}} & Q_{\mathrm{cal}}
\end{bmatrix}^{\mathsf T},
\label{eq:capability}
\end{equation}
where the entries represent task performance, generalisation, retention, reliability, efficiency, and calibration. This vector is illustrative rather than prescriptive. Each component must be translated into an operational metric with a defined scoring rule, task population, aggregation procedure, and direction of preference. For example, latency and cost are naturally minimisation metrics; they can be transformed into a higher-is-better score or handled by a separately specified upper confidence bound. The operational importance of these dimensions is also reflected in LLM-serving systems that optimise KV-cache memory use and the throughput--latency trade-off, including PagedAttention and Sarathi-Serve \citep{kwon2023pagedattention,agrawal2024sarathi}. The present theory assumes higher scores are better and scores are bounded in $[0,1]$.

The distinction between primary and protected metrics is a policy choice made before candidate evaluation. A metric can be primary in one update campaign and protected in another. If a metric is assigned both roles for the same candidate but with different comparators, it should be represented as two tests rather than silently combining the roles. This avoids ambiguity in the number of tests and in the family-wise error allocation.

\section{Evidence-Governed Adoption Rule}
For each primary objective $r\in\mathcal J$, let $\delta_r>0$ be the minimum practically meaningful improvement. For each protected capability $r\in\mathcal K$, let $\epsilon_r\geq0$ be the maximum tolerated degradation. These margins must be fixed before audit evaluation.

The population-level acceptance requirements are
\begin{align}
&\exists r\in\mathcal J:\quad \Delta^R_{j,r}\geq\delta_r,
\label{eq:primary}\\
&\forall r\in\mathcal K:\quad \Delta^R_{j,r}\geq-\epsilon_r,
\label{eq:protected}\\
&\forall m\in\mathcal G:\quad G_m(\widetilde A_j)=\mathrm{true},
\label{eq:hard}
\end{align}
where $\mathcal G$ is the set of mandatory governance checks. Examples include tool permissions, data-access restrictions, resource ceilings, and human-approval requirements. A high task score cannot compensate for failure of a mandatory constraint. Governance checks should be classified by enforcement mechanism. Deterministic checks can include allow-listed tool permissions, schema validation, network egress restrictions, resource ceilings, required approvals, and immutable audit logging. Human-review checks can cover changes with high impact or ambiguous consequences. Learned classifiers may assist triage, but should not be the sole enforcement mechanism for requirements whose violation is unacceptable. A check should specify its input, pass/fail semantics, version, owner, and failure action.

\begin{definition}[Evidence-governed candidate]
A candidate is evidence-governed if it is frozen before its audit sample is drawn, evaluated using fresh audit observations not used to construct or tune that candidate, and assessed using prespecified metrics, margins, and governance checks.
\end{definition}

\subsection{Evaluator Calibration}
Let $q^{E}_r(A,Z)$ denote the operational evaluator score and $q^{R}_r(A,Z)$ the score produced by a prespecified reference-evaluation procedure. The reference procedure may be a deterministic verifier, an adjudicated human protocol, or another independently governed measurement procedure. The target estimand in this subsection is the mean paired difference under the reference procedure, not an unobservable notion of ``true capability'' unless the reference procedure is itself a valid ground-truth measurement.

For candidate $j$, comparator $B_{j,r}$, and instance $Z$, define
\begin{align}
D^{E}_{j,r}(Z)&=q^{E}_r(\widetilde A_j,Z)-q^{E}_r(B_{j,r},Z),\\
D^{R}_{j,r}(Z)&=q^{R}_r(\widetilde A_j,Z)-q^{R}_r(B_{j,r},Z),\\
W_{j,r}(Z)&=D^{E}_{j,r}(Z)-D^{R}_{j,r}(Z).
\end{align}
Scores are in $[0,1]$, so $D^E,D^R\in[-1,1]$ and $W\in[-2,2]$. Write $\mu^E_{j,r}=\mathbb E[D^E_{j,r}\mid\mathcal H_j]$, $\Delta^R_{j,r}=\mathbb E[D^R_{j,r}\mid\mathcal H_j]$, and $b_{j,r}=\mathbb E[W_{j,r}\mid\mathcal H_j]$. Then $\mu^E_{j,r}-\Delta^R_{j,r}=b_{j,r}$.

Unlike a fixed, unexplained allowance, EGE estimates the discrepancy for each frozen candidate using an independent calibration sample drawn from the same declared metric distribution. This is intentionally demanding: a candidate-specific calibration set is needed because an update may change output style or behaviour in ways that affect evaluator agreement. A global calibration result can replace this design only if a separate uniform-generalisation argument covers the complete candidate class and adaptive selection process.

The calibration procedure does not eliminate construct-validity concerns. If the reference evaluator is systematically wrong, the resulting guarantee is only relative to that reference procedure. If human adjudication is used, the paper's target is the expected adjudicated score under the stated protocol; inter-rater disagreement and adjudication reliability must be reported. The reference procedure, rubric, blinding, version, and sampling frame must be fixed before confirmatory results are examined.

\subsection{Evidence Partitioning and Audit Integrity}
A practical implementation should distinguish at least three information partitions. The development partition is available to the candidate generator and may be used for prompt iteration, debugging, reflection, and search. The validation partition may be used to compare alternative candidate designs during development, but repeated adaptive use means it should not be treated as a final confirmatory sample. The audit partition is reserved for the frozen candidate and is revealed only after the candidate, comparator, metrics, margins, and analysis procedure have been committed.

The audit partition should be controlled as an operational resource. Access logs should record which candidate was evaluated, which audit instances were used, the evaluator version, the scoring rubric, and whether any audit result was exposed to the candidate generator. If an audit result is used to modify the candidate, that sample is no longer fresh for the modified candidate. The modified candidate must receive a new confirmatory sample or be evaluated with a method that explicitly accounts for adaptive reuse. This requirement is stronger than merely labelling a dataset as a test set: the information flow from audit results back into candidate construction must also be controlled.

Where tasks are clustered by user, session, domain, or environment, sampling individual rows independently may be inappropriate. The unit of sampling should reflect the dependence structure. For example, if several test cases arise from one user session, treating them as independent can understate uncertainty. A cluster-level design, a concentration inequality for dependent observations, or an appropriately conservative aggregation is then required. The Hoeffding result below is not automatically valid for arbitrary correlated task traces.

\section{Sequential Statistical Adoption}
\subsection{Sampling and Calibration Assumptions}
For each candidate $j$ and metric $r$, collect $n_{j,r}$ fresh confirmatory observations $d^E_{i,j,r}$ and $m_{j,r}$ independent calibration observations $w_{i,j,r}$. Define their means by
\begin{align}
\bar d^E_{j,r}&=\frac{1}{n_{j,r}}\sum_{i=1}^{n_{j,r}}d^E_{i,j,r},\label{eq:mean}\\
\bar w_{j,r}&=\frac{1}{m_{j,r}}\sum_{i=1}^{m_{j,r}}w_{i,j,r}.\label{eq:calmean}
\end{align}

\begin{assumption}[Conditional fresh sampling]
Conditional on the pre-audit history $\mathcal H_j$, candidate and comparator are frozen before either sample is drawn. Confirmatory observations are independent, lie in $[-1,1]$, and have common conditional mean $\mu^E_{j,r}$. Calibration observations are independent, lie in $[-2,2]$, and have common conditional mean $b_{j,r}$. The two samples are mutually independent conditional on $\mathcal H_j$, and neither is used to construct or tune the frozen candidate. Samples are drawn from the declared distribution for metric $r$. 
\label{ass:fresh}
\end{assumption}

\begin{assumption}[Finite prespecified metric family]
The primary and protected test sets are fixed and finite. Let $R=|\mathcal J|+|\mathcal K|$, counting a metric twice if it appears in both roles with different comparators or thresholds. Metric definitions, score ranges, distributions, margins, and reference-evaluation procedures are fixed before confirmatory outcomes are observed.
\label{ass:metrics}
\end{assumption}

\begin{assumption}[Reference-procedure validity]
For each test, the reference evaluator defines the target estimand $\Delta^R_{j,r}$. Any claim that this estimand represents an external capability relies on the validity of the reference procedure and its sampling frame. The theorem below does not establish that construct-validity claim.
\label{ass:reference}
\end{assumption}

\subsection{Evaluation and Calibration Bounds}
For an evaluation error allocation $e_{j,r}\in(0,1)$, Hoeffding's inequality gives
\begin{equation}
L^E_{j,r}=\bar d^E_{j,r}-\sqrt{\frac{2\ln(1/e_{j,r})}{n_{j,r}}},
\label{eq:eval-lcb}
\end{equation}
with $\Pr(L^E_{j,r}>\mu^E_{j,r}\mid\mathcal H_j)\leq e_{j,r}$.

For a calibration error allocation $c_{j,r}\in(0,1)$, the one-sided Hoeffding inequality for observations in an interval of width four gives
\begin{equation}
U^b_{j,r}=\bar w_{j,r}+\sqrt{\frac{8\ln(1/c_{j,r})}{m_{j,r}}},
\label{eq:bias-ucb}
\end{equation}
with $\Pr(b_{j,r}>U^b_{j,r}\mid\mathcal H_j)\leq c_{j,r}$.

Since $\Delta^R_{j,r}=\mu^E_{j,r}-b_{j,r}$, a lower confidence bound for the reference-procedure estimand is
\begin{equation}
L^R_{j,r}=L^E_{j,r}-U^b_{j,r}.
\label{eq:adjusted}
\end{equation}
On the joint event $L^E_{j,r}\leq\mu^E_{j,r}$ and $b_{j,r}\leq U^b_{j,r}$, we have
\[
L^R_{j,r}\leq\mu^E_{j,r}-U^b_{j,r}\leq\mu^E_{j,r}-b_{j,r}=\Delta^R_{j,r}.
\]
The calibration bound is one-sided because only optimistic evaluator discrepancy can make an evaluator-based lower bound overstate the reference-procedure effect. Negative discrepancy is not subtracted as if it were harmful bias; the resulting lower bound remains valid for the signed target estimand.

\subsection{Error Allocation Across Candidates and Tests}
Let $\alpha\in(0,1)$ be the global error budget. For global audit-event index $j=1,2,\ldots$ and test $r=1,\ldots,R$, allocate
\begin{equation}
e_{j,r}=c_{j,r}=\frac{3\alpha}{\pi^2j^2R}.
\label{eq:spend}
\end{equation}
Because $\sum_{j\geq1}j^{-2}=\pi^2/6$, the evaluation allocations sum to $\alpha/2$ and the calibration allocations also sum to $\alpha/2$. Therefore,
\begin{equation}
\sum_{j=1}^{\infty}\sum_{r=1}^{R}(e_{j,r}+c_{j,r})=\alpha.
\label{eq:sum}
\end{equation}

\begin{theorem}[Simultaneous evaluator-calibrated coverage]
Under Assumptions~\ref{ass:fresh}--\ref{ass:reference}, with the allocations in (\ref{eq:spend}),
\begin{equation}
\Pr\left\{L^R_{j,r}\leq\Delta^R_{j,r}\ \text{for every }j\geq1,\ r\in\{1,\ldots,R\}\right\}\geq1-\alpha.
\label{eq:coverage}
\end{equation}
No independence across candidate indices or metric tests is required for the final union-bound step.
\end{theorem}
\begin{proof}
For each fixed $(j,r)$, Assumption~\ref{ass:fresh} and Hoeffding's inequality imply that the evaluation-bound failure probability is at most $e_{j,r}$ and the calibration-bound failure probability is at most $c_{j,r}$. By the deterministic implication preceding the theorem, failure of $L^R_{j,r}\leq\Delta^R_{j,r}$ is contained in the union of those two events. Hence
\[
\Pr\{L^R_{j,r}>\Delta^R_{j,r}\}\leq e_{j,r}+c_{j,r}.
\]
A union bound over the countable candidate sequence and finite metric family yields
\[
\Pr\{\exists j,r:L^R_{j,r}>\Delta^R_{j,r}\}
\leq\sum_{j=1}^{\infty}\sum_{r=1}^{R}(e_{j,r}+c_{j,r})=\alpha.
\]
This proves the result. Adaptivity of candidate generation is permitted because the candidate is fixed conditional on $\mathcal H_j$ before its independent fresh evaluation and calibration samples are drawn. The index is a global audit-event index, allocated by the audit controller to candidate tests and reference-bridge tests alike; it cannot be reset after rejection, rollback, or a change in proposal strategy.
\end{proof}

\begin{remark}[What the theorem does not establish]
The theorem controls simultaneous statistical undercoverage relative to the reference-evaluation estimand. It does not prove that the reference evaluator measures the intended real-world capability, that the audit distribution matches deployment, or that an agent cannot cause harm through unmeasured actions. Those are separate validity and governance questions.
\end{remark}

\subsection{Audit Planning and a Conservative Power Condition}\label{sec:power}
Define the evaluation and calibration radii
\[
h^E_{j,r}=\sqrt{\frac{2\ln(1/e_{j,r})}{n_{j,r}}},\qquad
h^b_{j,r}=\sqrt{\frac{8\ln(1/c_{j,r})}{m_{j,r}}}.
\]
For a test with threshold $\tau_r$ and true reference-procedure effect $\Delta^R_{j,r}=\tau_r+g$ for $g>0$, the event $\bar d^E\geq\mu^E-h^E$ and $\bar w-b\leq h^b$ implies
\[
L^R_{j,r}\geq\Delta^R_{j,r}-h^E_{j,r}-2h^b_{j,r}.
\]
Consequently, on the joint event $\bar d^E_{j,r}\geq\mu^E_{j,r}-h^E_{j,r}$ and $\bar w_{j,r}-b_{j,r}\leq h^b_{j,r}$, the constructed lower bound obeys
\[
L^R_{j,r}\geq\Delta^R_{j,r}-2h^E_{j,r}-2h^b_{j,r}.
\]
The factor of two on the evaluation radius is necessary because the lower confidence bound itself subtracts one radius from the observed mean, while the event used to lower-bound that observed mean contributes another radius. Therefore, if $2h^E_{j,r}+2h^b_{j,r}\leq g$, the test passes with probability at least $1-e_{j,r}-c_{j,r}$. This is a sufficient, conservative power condition, not a necessary condition.

For any split parameter $\lambda\in(0,1)$, it is sufficient to choose
\begin{align}
n_{j,r}&\geq\frac{8\ln(1/e_{j,r})}{\lambda^2g^2},\label{eq:power-n}\\
m_{j,r}&\geq\frac{32\ln(1/c_{j,r})}{(1-\lambda)^2g^2}.\label{eq:power-m}
\end{align}
The parameter $\lambda$ allocates the available margin between confirmatory evaluation and calibration uncertainty. The evaluation bound above is four times the fixed-radius requirement in (\ref{eq:samplesize}) because the power argument must reserve margin both for the LCB construction and for downward sampling fluctuation. These bounds expose the cost of candidate-wise calibration: the required calibration sample can be substantial when the desired margin is small or the candidate index is large. They also show why a claim of practical efficiency requires empirical cost reporting rather than only a validity theorem.

\subsection{Statistical Acceptance Predicate}
The candidate passes the statistical gate if
\begin{align}
\Accept_j^{\mathrm{stat}}={}&
\left[\exists r\in\mathcal J:L^R_{j,r}\geq\delta_r\right]\nonumber\\
&\land\left[\forall r\in\mathcal K:
L^R_{j,r}\geq-\epsilon_r\right].
\label{eq:stataccept}
\end{align}
The lower-bound rule is intentionally conservative. A candidate does not pass merely because its observed mean is positive; the lower confidence bound must exceed the practical improvement margin. This distinguishes statistical evidence from effect-size relevance. A very small positive effect can be statistically distinguishable from zero yet still be operationally unimportant. Conversely, a candidate with a large observed effect but a small audit sample may remain inconclusive.

The adjusted lower bound can be read as the operational-evaluator sample mean minus the evaluation radius and the one-sided calibration upper bound. A primary test passes when $\bar d^E_{j,r}-h^E_{j,r}-U^b_{j,r}\geq\delta_r$. A protected test passes when $\bar d^E_{j,r}-h^E_{j,r}-U^b_{j,r}\geq-\epsilon_r$. This decomposition is useful for audit planning because it shows whether sampling uncertainty or evaluator discrepancy dominates an inconclusive decision.

\subsection{Interpretation of Inconclusive Outcomes}
The framework has three operationally distinct outcomes. A candidate can be accepted when every required statistical and governance condition passes. It can be rejected when a hard constraint fails or the evidence establishes that a required condition is not met. It can also remain inconclusive when the lower bound does not pass a threshold but the data do not establish a material regression. In a conservative deployment policy, both rejection and inconclusive outcomes retain the incumbent; however, the distinction should be preserved in logs because it matters for subsequent candidate development and audit-cost analysis. Additional sampling is permissible only if the sampling plan was prespecified or if the monitoring procedure is sequentially valid.

The complete offline acceptance predicate is
\begin{equation}
\Accept_j=\Accept_j^{\mathrm{stat}}\land
\bigwedge_{m\in\mathcal G}
\{G_m(\widetilde A_j)=\mathrm{true}\}.
\label{eq:accept}
\end{equation}

\begin{corollary}[Family-wise control of false statistical acceptance]
Under the theorem's assumptions and the evaluator-calibration procedure, the probability that any candidate in the audit campaign is statistically accepted while failing at least one population requirement in (\ref{eq:primary})--(\ref{eq:protected}) is at most $\alpha$.
\end{corollary}
\begin{proof}
On the simultaneous-coverage event, every adjusted lower bound is no greater than its target population difference. If all primary objectives fail their respective meaningful-improvement thresholds, no primary lower bound can pass. If any protected capability violates its non-inferiority margin, its lower bound cannot pass. Thus a population-invalid candidate cannot pass the statistical gate on the simultaneous-coverage event. The complement has probability at most $\alpha$.
\end{proof}

\section{Reference-Anchored Capability Stability and Reference Lineage}
A fixed reference prevents the protected baseline from moving after every accepted update. If the reference is eventually replaced, however, a new baseline can reintroduce cumulative drift unless the replacement is explicitly compared with its predecessor. EGE therefore treats reference replacement as a separately audited governance event.

\begin{proposition}[Fixed-reference non-inferiority]
On the simultaneous-coverage event in Theorem~1, every candidate that passes all protected-capability tests in a fixed-reference epoch satisfies
\[
\Delta^R_{j,r}(\widetilde A_j,A_{\mathrm{ref}})\geq-\epsilon_r,
\qquad r\in\mathcal K.
\]
\end{proposition}
\begin{proof}
For every protected test, simultaneous coverage gives $L^R_{j,r}\leq\Delta^R_{j,r}$. Acceptance requires $L^R_{j,r}\geq-\epsilon_r$. Combining the inequalities proves the claim.
\end{proof}

\subsection{Reference Replacement and Bridge Tests}
Let $A^{(0)},A^{(1)},\ldots,A^{(K)}$ denote a sequence of approved references. At transition $k$, a bridge test compares the proposed new reference $A^{(k+1)}$ with the previous reference $A^{(k)}$ on the same protected metric and target distribution. Let its target difference be
\[
B_{k,r}=Q^R_r(A^{(k+1)})-Q^R_r(A^{(k)}).
\]
The bridge test must use fresh evaluation and calibration samples and receive its own error allocation within the global budget. In implementation, every candidate test and every bridge test receives a unique, non-reusable index from one global audit controller; the summable allocation in (\ref{eq:spend}) is applied to this global event index. Reference transitions therefore cannot reuse error budget already spent on candidate tests. Let $L^{\mathrm{bridge}}_{k,r}$ be its evaluator-calibrated lower confidence bound. The replacement is permitted only if $L^{\mathrm{bridge}}_{k,r}\geq-\rho_{k,r}$, where $\rho_{k,r}\geq0$ is a prespecified bridge tolerance. A reference change is therefore not an administrative reset: it consumes statistical evidence and is recorded as a distinct lineage transition.

\begin{theorem}[Cumulative capability bound along a reference lineage]
Assume all candidate tests and all bridge tests are included in one simultaneous-coverage family with total error budget $\alpha$. Assume every committed configuration $A$ in epoch $K$ passes its protected-capability test relative to $A^{(K)}$ with tolerance $\epsilon_{K,r}$, and every accepted reference transition satisfies $B_{k,r}\geq-\rho_{k,r}$. Then, on the simultaneous-coverage event,
\[
Q^R_r(A)-Q^R_r(A^{(0)})\geq-
\left(\epsilon_{K,r}+\sum_{k=0}^{K-1}\rho_{k,r}\right).
\]
Thus the maximum allowed degradation relative to the original reference is explicit and accumulates only through declared epoch tolerances.
\end{theorem}
\begin{proof}
By the candidate's protected-capability acceptance condition and simultaneous coverage,
$Q^R_r(A)-Q^R_r(A^{(K)})\geq-\epsilon_{K,r}$. For each bridge transition, simultaneous coverage and its acceptance condition give
$Q^R_r(A^{(k+1)})-Q^R_r(A^{(k)})\geq-\rho_{k,r}$. Add these inequalities and telescope the intermediate reference terms:
\[
Q^R_r(A)-Q^R_r(A^{(0)})
=Q^R_r(A)-Q^R_r(A^{(K)})
+\sum_{k=0}^{K-1}\left[Q^R_r(A^{(k+1)})-Q^R_r(A^{(k)})\right].
\]
The claimed lower bound follows.
\end{proof}

The result requires the same metric definition and target distribution across all bridge comparisons. If the target distribution changes, the terms are not directly additive without an additional transport or standardisation argument. The bound is also only as meaningful as the reference-evaluation procedure. It formalises reference lineage; it does not prove that the initial reference is optimal or that the selected protected metrics exhaust all relevant capabilities.

\section{Evidence-Governed Deployment}
Offline statistical acceptance does not imply safe behaviour under production load or on unrepresented tasks. EGE therefore separates statistical acceptance from persistent commitment.

\subsection{Two-Phase Commitment}
The operational state machine has four states: (1) \emph{Candidate}, where a proposed modification is frozen; (2) \emph{Audit-passed}, where statistical and governance requirements pass; (3) \emph{Canary}, where the candidate runs under bounded traffic, permissions, workload, or execution scope; and (4) \emph{Committed}, where it becomes the persistent configuration after prespecified canary criteria pass.

A candidate failing offline evaluation is rejected. A candidate failing canary criteria is contained and the incumbent configuration is restored. Configuration rollback does not reverse irreversible external side effects; systems with consequential tool access need transaction controls and compensating-action procedures.

The canary scope can be limited along several dimensions: the fraction of eligible requests routed to the candidate, the set of users or tenants exposed, the classes of tools enabled, the maximum number of tool calls, the permitted data domains, the execution time, and the maximum resource budget. The appropriate controls depend on the agent's action space. For a read-only summarisation agent, a traffic-split canary may be sufficient. For an agent that can modify records or trigger external workflows, a shadow mode or approval-gated mode may be more appropriate than live execution.

The canary must have an explicit stop condition and an explicit authority to stop. A useful operational record includes the candidate hash, incumbent hash, canary start and end times, exposure volume, metric values, alert events, rollback reason, and any compensating actions. The candidate should not be permitted to change itself during the canary; otherwise, the object being monitored is no longer the object that passed the offline audit. If adaptation during canary operation is required, each resulting version should be treated as a new candidate or covered by a separately justified online-learning protocol.

\subsection{Operational Monitoring}
Canary criteria should be application-specific and fixed before deployment. They may include task success, tool-error rate, latency, resource consumption, policy violations, escalation rate, and user-impact indicators. Repeated monitoring requires sequentially valid procedures or deterministic operational limits; a fixed-sample confidence interval must not be repeatedly inspected as if it were time-uniform.

\subsection{Risk-Tiered Deployment}
A practical deployment controller can assign candidate changes to risk tiers before evaluation. A low-impact change might alter wording or reorder a non-critical retrieval preference while preserving permissions and tool interfaces. A medium-impact change might modify memory retrieval, task routing, or multi-step workflow logic. A high-impact change might modify access to sensitive data, external-action permissions, financial or operational workflows, or safety-relevant control logic. These examples are illustrative; the organisation must define its own classification.

The risk tier should determine the required canary scope, duration, approval level, and rollback authority. It should not change the statistical meaning of the audit results after they have been observed. Higher-risk changes may require stricter margins, more protected metrics, deterministic checks, or mandatory human approval. Such requirements should be encoded before the candidate is evaluated. This preserves the separation between evidence and policy while allowing governance controls to scale with consequence.

\subsection{Rollback and Recovery}
Rollback should be implemented as a tested recovery procedure rather than an informal instruction to restore a prior prompt. A complete recovery package may include the prior configuration, memory-schema compatibility information, tool and API versions, feature flags, routing state, and any migration steps required to restore service. If a candidate changes persistent memory, rollback may require restoring a snapshot or applying a reversible migration. If it writes to external systems, recovery may require compensating transactions and an incident process. Some side effects cannot be reversed; the system should identify those actions before enabling the candidate and apply stricter controls accordingly.

\begin{figure*}[!t]
\centering
\resizebox{\linewidth}{!}{%
\begin{tikzpicture}[
  >=Latex,
  font=\footnotesize,
  node distance=8mm and 10mm,
  process/.style={
    draw=black!75, rounded corners=2pt, line width=0.7pt,
    align=center, text width=29mm, minimum height=13mm,
    inner sep=4pt
  },
  source/.style={process, fill=blue!5},
  evidence/.style={process, fill=green!6},
  deployment/.style={process, fill=orange!8},
  reject/.style={process, fill=red!5},
  flow/.style={-{Latex[length=2mm]}, line width=0.75pt, draw=black!80},
  flowlabel/.style={font=\scriptsize, fill=white, inner sep=1.5pt, text=black!85}
]
% Evidence pipeline: one unambiguous left-to-right sequence.
\node[source] (inc) at (0,0)
  {\textbf{Incumbent and experience}\\$I_j,\;\mathcal E_j$};
\node[source] (candidate) at (4.0,0)
  {\textbf{Candidate generation}\\Freeze $\widetilde A_j$};
\node[evidence] (audit) at (8.0,0)
  {\textbf{Fresh audit}\\Paired metrics and LCBs};
\node[evidence] (gate) at (12.0,0)
  {\textbf{Governance gate}\\Margins and hard checks};
\draw[flow] (inc.east) -- (candidate.west);
\draw[flow] (candidate.east) -- (audit.west);
\draw[flow] (audit.east) -- (gate.west);

% Outcomes are placed on a separate row to keep decision paths apart.
\node[reject] (reject) at (4.0,-2.65)
  {\textbf{Reject or roll back}\\Retain incumbent $I_j$};
\node[deployment] (canary) at (12.0,-2.65)
  {\textbf{Bounded canary}\\Operational monitoring};
\node[deployment] (commit) at (16.0,-2.65)
  {\textbf{Persistent commit}\\$I_{j+1}=\widetilde A_j$};

% A vertical pass path and a separate, non-crossing failure path.
\draw[flow] (gate.south) -- node[flowlabel, right, xshift=1mm] {Pass} (canary.north);
\draw[flow] (gate.south west) -- ++(-0.45,-0.45) -- (4.0,-0.9)
  -- (reject.north);
\node[flowlabel] at (7.5,-1.12) {Offline checks fail};

% Canary success proceeds to commitment; failure returns below the boxes.
\draw[flow] (canary.east) -- node[flowlabel, above] {Canary passes} (commit.west);
\draw[flow] (canary.south) -- ++(0,-0.62) -| (reject.south);
\node[flowlabel] at (8.0,-3.55) {Canary fails};
\end{tikzpicture}%
}
\caption{Evidence-Governed Evolution separates candidate generation, fresh statistical evaluation, mandatory governance checks, and bounded deployment. Offline failure leads to rejection; only a candidate that passes both offline checks and canary monitoring is persistently committed.}
\label{fig:ege}
\end{figure*}
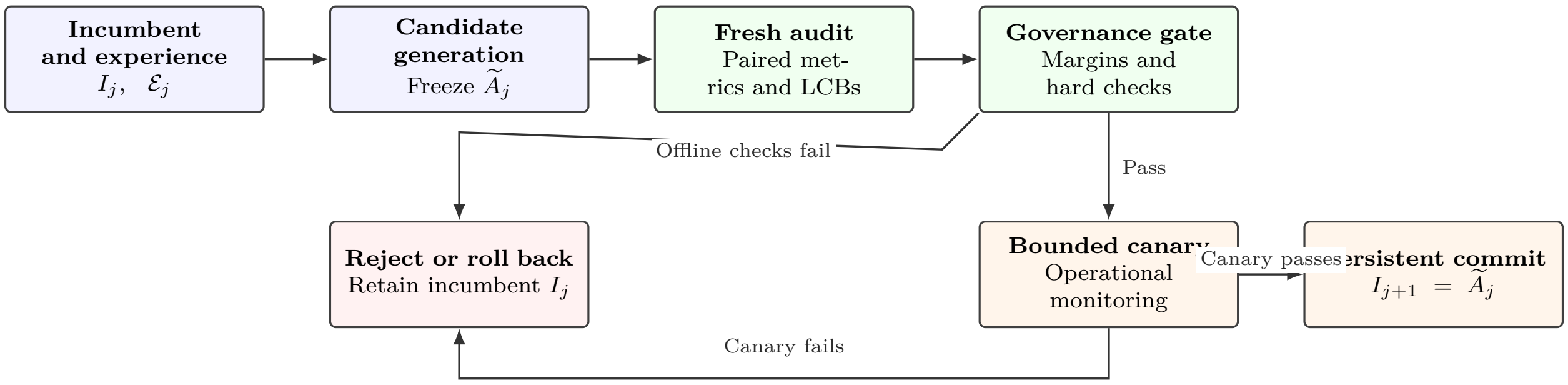

\begin{algorithm}[!t]
\caption{Evidence-Governed Evolution}
\label{alg:ege}
\begin{algorithmic}[1]
\Require Current incumbent $I_j$, reference $A_{\mathrm{ref}}$, experience $\mathcal E_j$
\Require Margins $\delta,\epsilon$, error budget $\alpha$, metric count $R$
\State Generate $\widetilde A_j\gets\mathcal U(I_j,\mathcal E_j)$
\State Freeze candidate; do not tune it using confirmatory audit outcomes
\State For each metric $r$, draw independent fresh confirmatory and calibration samples
\State Compute $e_{j,r},c_{j,r}$ using (\ref{eq:spend})
\State Compute evaluator LCB and calibration UCB using (\ref{eq:eval-lcb})--(\ref{eq:bias-ucb})
\State Compute reference-procedure LCB $L^R_{j,r}$ using (\ref{eq:adjusted})
\State Evaluate primary and protected-capability tests in (\ref{eq:stataccept})
\State Evaluate all mandatory governance checks
\If{any required condition fails}
  \State \Return Reject candidate; retain $I_j$
\EndIf
\State Deploy candidate in bounded canary scope
\If{all prespecified canary criteria pass}
  \State $I_{j+1}\gets\widetilde A_j$
  \State \Return Commit $I_{j+1}$
\Else
  \State Contain candidate; restore $I_j$
  \State Execute compensating actions where applicable
  \State \Return Roll back; retain $I_j$
\EndIf
\end{algorithmic}
\end{algorithm}

\section{Evaluation}
This section reports one executed, fully specified Monte Carlo stress test and defines the additional real-agent evaluation required to assess operational usefulness. The Monte Carlo study is synthetic by construction: it tests a statistical failure mode under known data-generating assumptions. It is not evidence that EGE improves the behaviour of an implemented autonomous agent.

\subsection{Monte Carlo sensitivity study: evaluator bias and audit size}
\paragraph{Purpose and design.} We ran a targeted Monte Carlo sensitivity study to examine two distinct properties of the proposed gate: (i) whether candidate-specific calibration prevents harmful acceptance when the operational evaluator is optimistic, and (ii) how much acceptance power is lost when calibration and audit samples are small. This is a synthetic statistical experiment, not an autonomous-agent benchmark. Each condition comprises 3,000 independent campaigns with 10 candidates per campaign. Each candidate is beneficial or harmful with probability one half. The reference-procedure primary effect is $+0.25$ for beneficial candidates and $-0.10$ for harmful candidates; the protected effect is zero. The primary threshold is $\delta=0.05$, the protected non-inferiority margin is $\epsilon=0.20$, and the global error budget is $\alpha=0.05$. The operational evaluator's primary-score optimism is varied over $b\in\{0,0.10,0.25\}$; protected-metric bias is zero. Confirmatory paired differences are sampled from $\{-1,+1\}$ with the specified means, while calibration discrepancies are sampled from $\{-2,+2\}$ with the specified mean. The confirmatory and calibration samples are independent. We vary both sample sizes together over $n=m\in\{5{,}000,20{,}000,80{,}000\}$ per candidate and metric. The candidate-wise evaluation and calibration allocations are each $3\alpha/(\pi^2j^2R)$, so the two families jointly consume the global error budget. The random seed is 20261112.

\paragraph{Comparators and outcomes.} The calibrated EGE comparator uses the evaluation lower confidence bound and the independent one-sided calibration upper confidence bound. The evaluator-naive comparator uses the same evaluation lower confidence bound but omits calibration. A campaign-level harmful-acceptance event occurs if at least one harmful candidate passes the primary and protected tests. Beneficial-candidate acceptance is the proportion of beneficial candidates accepted. The table reports observed campaign rates and beneficial-candidate acceptance frequencies. For zero observed campaign events, the one-sided exact 95\% binomial upper confidence limit is reported; for nonzero event rates, two-sided exact 95\% Clopper--Pearson intervals are used in the supplementary JSON.

\begin{table}[t]
\caption{Monte Carlo sensitivity to evaluator optimism and sample size (3,000 campaigns per condition; 10 candidates per campaign).}
\label{tab:mc-sensitivity}
\centering
\small
\begin{tabular}{rrrrrr}
\toprule
$n=m$ & Bias $b$ & EGE harmful campaigns & Naive harmful campaigns & EGE power & Naive power \\
\midrule
5,000 & 0.00 & 0/3,000 & 0/3,000 & 52.6\% & 100.0\% \\
5,000 & 0.10 & 0/3,000 & 0/3,000 & 76.2\% & 100.0\% \\
5,000 & 0.25 & 0/3,000 & 2,997/3,000 & 65.7\% & 100.0\% \\
20,000 & 0.00 & 0/3,000 & 0/3,000 & 100.0\% & 100.0\% \\
20,000 & 0.10 & 0/3,000 & 0/3,000 & 100.0\% & 100.0\% \\
20,000 & 0.25 & 0/3,000 & 2,997/3,000 & 100.0\% & 100.0\% \\
80,000 & 0.00 & 0/3,000 & 0/3,000 & 100.0\% & 100.0\% \\
80,000 & 0.10 & 0/3,000 & 0/3,000 & 100.0\% & 100.0\% \\
80,000 & 0.25 & 0/3,000 & 2,997/3,000 & 100.0\% & 100.0\% \\
\bottomrule
\end{tabular}
\end{table}

For the calibrated rule, no harmful campaign acceptance was observed in any of the nine correctly specified conditions. The corresponding one-sided exact 95\% upper confidence limit is approximately $9.98\times10^{-4}$ for each condition; this Monte Carlo limit is not the theorem's $\alpha$ guarantee. Under $b=0.25$, the evaluator-naive rule accepted at least one harmful candidate in 2,997 of 3,000 campaigns (99.9\%). Under $b=0$ and $b=0.10$, it recorded no harmful campaign acceptance in this particular design. The contrast is therefore specific to the selected bias/effect combinations and should not be read as a general ranking of methods under all evaluator regimes.

The calibrated method's beneficial-candidate acceptance frequency was 52.6--76.2\% at 5,000 observations per sample, then 100\% at 20,000 and 80,000 observations. This illustrates a real trade-off: calibration can block optimistic false gains but can also make the gate inconclusive when audit resources are insufficient. The 100\% frequencies arise from deliberately separated effects and are not a universal power guarantee. Full campaign-level counts, confidence intervals, and seed are included in the reproducibility files.

\subsection{Calibration-distribution mismatch stress test}
The guarantee requires the calibration and confirmatory samples to represent the same declared metric distribution. To illustrate why this assumption matters, we ran a deliberately misspecified condition with 3,000 campaigns, 10 harmful candidates per campaign, $n=m=100{,}000$, reference-procedure primary effect $-0.10$, actual audit-evaluator optimism $+0.25$, but calibration-sample optimism only $+0.05$. All other settings and the nominal allocation were unchanged. The calibrated procedure accepted at least one harmful candidate in all 3,000 campaigns (100\%). This is not a counterexample to the theorem: the calibration sample is not drawn from the same discrepancy distribution as the audit sample, so the stated assumptions are violated. It is a diagnostic showing that a formal confidence bound cannot compensate for a nonrepresentative calibration design.

\begin{table}[t]
\caption{Calibration-distribution mismatch stress test (assumption deliberately violated).}
\label{tab:cal-mismatch}
\centering
\small
\begin{tabular}{lrrr}
\toprule
Actual audit bias & Calibration bias & Harmful-acceptance campaigns & Observed rate \\
\midrule
$+0.25$ & $+0.05$ & 3,000/3,000 & 100.0\% \\
\bottomrule
\end{tabular}
\end{table}

The exact two-sided 95\% Clopper--Pearson interval for the observed 3,000/3,000 rate is approximately $[0.9988,1.0000]$. This interval quantifies Monte Carlo uncertainty only; it does not turn the misspecified experiment into a valid test of EGE's theorem.

\begin{table}[t]
\caption{Illustrative audit and calibration sample-size requirements under the corrected sufficient-power condition.}
\label{tab:sample-planning}
\centering
\small
\begin{tabular}{rrrrr}
\toprule
Global event index $j$ & Per-test $e=c$ & $n$ minimum & $m$ minimum & Total \\
\midrule
1 & $5.07\times10^{-3}$ & 16,913 & 67,651 & 84,564 \\
10 & $5.07\times10^{-5}$ & 31,650 & 126,597 & 158,247 \\
100 & $5.07\times10^{-7}$ & 46,386 & 185,543 & 231,929 \\
\bottomrule
\end{tabular}
\end{table}

The planning table uses $\alpha=0.05$, $R=3$, target gap $g=0.10$, and split parameter $\lambda=0.5$. Counts are rounded upward from the sufficient conditions in Section~\ref{sec:power}; they are per metric test and per candidate/bridge event. The calibration requirement is larger because the discrepancy variable has range width four, compared with width two for the paired evaluation difference. The growth with $j$ reflects the summable allocation and makes the audit-cost implications explicit.

\subsection{Required Real-Agent Evaluation Protocol}
A real-agent study remains necessary before submission to a venue expecting evidence of practical utility. The study should use a reproducible agent whose editable harness is limited to declared components such as prompts, retrieval settings, tool-routing rules, or workflow configuration. Candidate generation must be separated from audit and reference-evaluation services. Each candidate should be frozen and content-hashed before independent audit and calibration samples are drawn.

\paragraph{Tasks and metrics.} Use a task suite with deterministic verifiers wherever possible; benchmark design and red-teaming studies can inform breadth and adversarial coverage \citep{hendrycks2021measuring,perez2022redteaming}. Include at least one primary objective and multiple protected capabilities chosen before the experiment. Candidate and comparator should be evaluated on the same task instances when pairing is valid. For stochastic agents, record model version, decoding settings, tool versions, seeds, timeouts, and retry rules. If tasks are clustered by user/session or share state, use cluster-level sampling or a concentration result appropriate to the dependence structure. Do not apply the independent-observation Hoeffding bound to correlated traces without justification.

\paragraph{Candidate generation and baselines.} Use the same candidate proposal stream for all adoption strategies where feasible. Compare (i) greedy observed-score acceptance, (ii) fixed observed-mean thresholds, (iii) confidence-bound acceptance without evaluator calibration, (iv) calibrated EGE without reference protection, (v) calibrated EGE with incumbent-only protection, and (vi) full EGE. Report candidate counts, accepted candidates, harmful acceptances, false rejections, inconclusive decisions, audit observations, calibration observations, elapsed time, and canary exposure.

\paragraph{Ablations and stress conditions.} Vary candidate index, audit size, calibration size, evaluator optimism, evaluator drift, protected-capability regression, and task-distribution shift. Include an evaluator-manipulation condition in which candidate changes alter output style or structure while leaving task success unchanged. Report evaluator/reference disagreement by candidate class and style. Use blinded human adjudication or deterministic verifiers as the reference procedure; if human ratings are used, report the rubric, rater training, agreement, adjudication, and sampling frame. The reference procedure must not be described as ground truth unless that claim is justified.

\paragraph{Statistical reporting.} Report campaign-level false-acceptance rates with exact or otherwise justified uncertainty intervals, acceptance power under prespecified effect sizes, per-capability regression rates, and audit cost. Report simulation seeds and number of repetitions. For agent experiments, report uncertainty across independent update episodes and avoid treating multiple tasks from one shared history as independent episodes. Separate offline false acceptance from canary failure: the former is controlled only under the theorem's assumptions, whereas the latter depends on deployment representativeness and monitoring design.

\paragraph{Canary study.} A candidate that passes offline evaluation should enter a bounded, immutable canary. Predeclare exposure limits, permitted tools, monitoring metrics, stop rules, and rollback authority. If canary metrics are inspected repeatedly, use a sequentially valid procedure or deterministic stop thresholds; do not repeatedly inspect a fixed-sample interval as if it were time-uniform. Record irreversible side effects and compensating actions. The canary stage must not be described as statistically guaranteed by the offline theorem.

\subsection{Reproducibility and Availability}
The supplementary simulation script specifies the random seed, data-generating process, allocation, sample sizes, thresholds, and outcome calculations. A real-agent reproduction package should additionally include candidate hashes, incumbent and reference versions, task provenance, evaluator prompts and versions, calibration records, raw aggregate scores, exclusion rules, and canary logs. Privacy- or security-sensitive task content may be withheld, but the paper should explain the resulting reproducibility limitation. No real-agent code or dataset is claimed to accompany this version.

\section{Discussion}
\subsection{Generation Is Not Adoption}
Candidate generators optimise the production of plausible modifications. Adoption mechanisms decide whether evidence justifies changing persistent state. If the same agent proposes, evaluates, and endorses changes using overlapping evidence, a self-confirming loop can arise. EGE makes evidence separation explicit without requiring human approval for every low-impact update. Human review remains appropriate for high-impact changes or uncertain evaluators. This separation also clarifies accountability: the candidate generator is responsible for proposing a change, the evaluation service is responsible for producing a measurement under a fixed protocol, and the governance authority is responsible for defining which risks and trade-offs are acceptable. These responsibilities may be implemented by separate services or teams, but they should remain distinguishable in logs and incident reviews.

\subsection{Evaluator Validity}
Fresh audit data do not guarantee evaluator correctness. The candidate-specific calibration construction accounts for sampling uncertainty in the observed discrepancy, but it does not repair an invalid reference construct or a calibration sample drawn from the wrong population. Candidate-specific calibration is expensive and may itself be vulnerable if the reference procedure is not blinded or independent. A model-based judge may favour a candidate's style or exploit a scoring rubric. Deterministic verifiers and blinded human adjudication are preferable where feasible. For subjective tasks, inter-rater agreement and construct validity matter as much as nominal sample size. A numerically precise interval around a systematically invalid construct is not evidence that the intended capability has been measured.

\subsection{Distribution Shift and Unmeasured Failure}
The statistical result concerns the declared distribution $\mathcal D_r$. It does not automatically transfer to shifted deployment tasks. Monitoring, re-audit, and suspension of automatic adoption may be needed when representativeness deteriorates. No finite metric set can guarantee preservation of capabilities that are not measured. Distribution monitoring should therefore be treated as a trigger for reassessment, not as proof that the original guarantees remain valid. A shift detector can itself produce false alarms or miss subtle changes, so its thresholds and response policy should be documented. When a material shift is detected, the organisation may suspend automatic adoption, collect a new representative audit set, revise the target distributions, or require additional human review. Any change to the reference distribution or protected metric definitions should be versioned.

\subsection{Conservatism and Cost}
The allocation in (\ref{eq:spend}) is intentionally conservative: later candidate indices receive smaller per-test error levels. This can require larger samples. Alternatives include a finite prespecified candidate horizon, confidence sequences, or more efficient concentration bounds. Any replacement must preserve the relevant family-wise guarantee under its own assumptions. There is also a distinction between the error budget and the desired power of the procedure. The alpha-spending schedule controls false statistical acceptance, but it does not guarantee that beneficial candidates will be accepted. Later candidates receive smaller error allocations, which can increase the required sample size and the probability of an inconclusive outcome at a fixed sample size. A deployment study should therefore report both error control and power or acceptance probability under specified effect sizes. The system owner must choose an acceptable balance between audit cost, decision delay, and missed improvements.

\section{Limitations and Future Work}
The guarantee assumes bounded paired observations and conditionally independent fresh confirmatory and calibration samples. Real agent tasks may be dependent, nonstationary, or affected by shared infrastructure. The reference-evaluation procedure must be sufficiently valid for the intended construct; the theorem only controls discrepancy relative to that procedure. Candidate-specific calibration is sample-intensive, and the Monte Carlo study is a stylised stress test rather than a real-agent evaluation. Extensions to clustered sampling, martingale differences, nonstationary evaluation, and efficient uniform calibration over candidate classes are important. A reference can become obsolete; the bridge-test result controls only declared metrics on a common distribution. Configuration rollback cannot undo external actions. The immediate empirical priority is to implement the protocol in a representative agent system and measure both audit cost and operational outcomes.

A further limitation concerns metric multiplicity within a capability. A single aggregate score can conceal subgroup regressions, while a large collection of subgroup metrics increases the number of simultaneous tests and therefore the required error allocation. Hierarchical testing or structured metric families may improve efficiency, but they require a prespecified hierarchy and a valid multiplicity procedure. Another limitation is that bounded normalised scores can discard information about severity. For example, a rare catastrophic event may be poorly represented by an average score in $[0,1]$. Such outcomes should be represented by separate hard constraints, tail-risk measures, or event-rate requirements rather than being diluted in an average capability score.

The framework also assumes that the comparator can be evaluated on the same instance as the candidate. This paired design is attractive because it removes some task-to-task variation, but it may not be feasible when candidate changes alter task eligibility, tool availability, or execution pathways. In such cases, the experiment may require matched cohorts, randomised assignment, or a different estimand. The resulting inference must reflect the actual design; the paired Hoeffding formula should not be applied mechanically.

Finally, EGE is a governance framework, not a complete agent security architecture. It does not by itself prevent prompt injection, data exfiltration, supply-chain compromise, malicious tool responses, or failures in the underlying model. Permission isolation, secure execution, logging, identity management, and incident response remain necessary. The contribution is narrower: EGE specifies a statistically explicit gate for persistent adoption and a deployment protocol that complements those controls.

\section{Conclusion}
This paper developed Evidence-Governed Evolution for the persistent adoption of self-generated modifications to autonomous-agent configurations. The framework separates candidate generation from confirmatory adoption, requires fresh paired evaluation, calibrates the operational evaluator against an independently governed reference procedure, protects declared capabilities relative to a versioned reference, and places accepted candidates behind bounded canary deployment. Under explicit conditional sampling assumptions, the evaluator-calibrated lower bounds provide simultaneous coverage over a countable sequence of adaptively generated candidates. A reference-lineage result makes the cost of replacing a protected baseline explicit, while an audit-planning inequality relates the effect-size margin to confirmatory and calibration sample requirements.

The Monte Carlo stress test demonstrates one narrow point: candidate-dependent evaluator optimism can make an evaluator-naive gate accept harmful updates, while candidate-specific calibration can prevent those acceptances in a stylised setting. It does not establish that EGE improves real-agent performance, is operationally efficient across realistic workloads, or remains valid under arbitrary dependence and distribution shift. Those questions require a controlled agent-task study and careful validation of the reference evaluator. EGE should therefore be viewed as a formal adoption protocol with testable assumptions, not as a proof of universal agent safety. The central principle remains that generating a plausible update is not evidence that it should become persistent system state.

\appendix
\section{Hoeffding Bound Derivation}
For independent $X_i\in[-1,1]$ with common mean $\mu$, Hoeffding's inequality gives
\[
\Prob(\bar X-\mu\geq u)\leq
\exp\left(-\frac{2nu^2}{(1-(-1))^2}\right)
=\exp(-nu^2/2).
\]
Setting this upper bound equal to $a$ yields
$u=\sqrt{2\ln(1/a)/n}$, and hence
\[
\Prob\left\{\mu\geq\bar X-
\sqrt{\frac{2\ln(1/a)}{n}}\right\}\geq1-a.
\]
This is the evaluation radius used in (\ref{eq:eval-lcb}); variance-sensitive alternatives such as empirical Bernstein bounds can be tighter when their assumptions are appropriate \citep{maurer2009empirical,boucheron2013concentration}.

\section{Calibration-Bound Derivation}
For independent calibration observations $W_i\in[-2,2]$ with mean $b$, the one-sided Hoeffding inequality gives
\[
\Pr(b-\bar W\geq u)\leq
\exp\left(-\frac{2mu^2}{(2-(-2))^2}\right)
=\exp(-mu^2/8).
\]
Setting the right-hand side to $c$ yields $u=\sqrt{8\ln(1/c)/m}$. Therefore $\Pr(b\leq\bar W+u)\geq1-c$, which establishes the one-sided calibration upper bound in (\ref{eq:bias-ucb}).

\section{Evaluation and Calibration Sample-Size Bounds}
To make the evaluation radius no greater than $h_E>0$, it is sufficient that
\begin{equation}
n_{j,r}\geq\frac{2\ln(1/e_{j,r})}{h_E^2}.
\label{eq:samplesize}
\end{equation}
To make the calibration radius no greater than $h_b>0$, it is sufficient that
\[
m_{j,r}\geq\frac{8\ln(1/c_{j,r})}{h_b^2}.
\]
For a target-effect gap $g$, the sufficient condition $2h_E+2h_b\leq g$ yields the acceptance-probability bound described in Section~\ref{sec:power}. These are conservative worst-case bounds; variance-adaptive alternatives require their own validity arguments.

\section{Notation and Implementation Checklist}
For clarity, Table~\ref{tab:notation} summarises the principal symbols used in the framework. The table is intended as an implementation aid; the formal definitions in the main text take precedence if a symbol is used in a specialised context.
\begin{table}[!t]
\caption{Principal notation used in EGE}
\label{tab:notation}
\centering
\footnotesize
\begin{tabularx}{\linewidth}{>{\raggedright\arraybackslash}p{0.19\linewidth} X}
\toprule
Symbol & Meaning \\
\midrule
$I_j$ & Incumbent configuration before candidate $j$ \\
$\widetilde A_j$ & Frozen candidate configuration \\
$A_{\mathrm{ref}}$ & Versioned stable reference configuration \\
$\mathcal E_j$ & Experience available to candidate generation \\
$\mathcal J$ & Prespecified primary-objective test set \\
$\mathcal K$ & Prespecified protected-capability test set \\
$\mathcal G$ & Mandatory governance checks \\
$\mathcal D_r$ & Evaluation distribution for metric $r$ \\
$\Delta^R_{j,r}$ & Reference-procedure population difference for candidate $j$ and metric $r$ \\
$b_{j,r}$ & Mean operational-to-reference evaluator discrepancy \\
$U^b_{j,r}$ & One-sided calibration upper confidence bound on $b_{j,r}$ \\
$e_{j,r},c_{j,r}$ & Evaluation and calibration error allocations \\
$m_{j,r}$ & Number of calibration observations \\
$\delta_r$ & Minimum meaningful primary improvement \\
$\epsilon_r$ & Maximum tolerated protected-capability degradation \\
$\alpha$ & Global family-wise error budget \\
$n_{j,r}$ & Number of confirmatory paired observations \\
$L^R_{j,r}$ & Evaluator-calibrated lower confidence bound for the reference estimand \\
\bottomrule
\end{tabularx}
\end{table}

A reference implementation should verify the following before enabling automatic commitment:
\begin{enumerate}
\item The candidate and comparator identifiers are immutable for the duration of the audit.
\item The candidate index is allocated by a persistent audit controller and cannot be reset after rejection.
\item The primary and protected metric families, their comparators, thresholds, and evaluation distributions are versioned.
\item The audit sample is fresh for the frozen candidate and is not fed back into its development loop.
\item The sampling unit and dependence assumptions match the task-generation process.
\item The score range is enforced or justified; observations outside the assumed interval are treated as protocol failures.
\item The candidate-specific calibration sample is independent of the confirmatory sample and is drawn from the declared evaluation distribution.
\item Every hard governance check has a deterministic or authorised review procedure and a defined failure action.
\item Canary exposure is bounded, monitored, and stopped by a controller independent of the candidate.
\item Rollback, state restoration, and compensating actions are tested before deployment.
\end{enumerate}

This checklist does not add new statistical guarantees. It translates the assumptions and protocol requirements into auditable implementation conditions.

\section{Worked Illustrative Decision}
Consider one primary test and two protected tests for candidate $j=1$, with $R=3$ and global error budget $\alpha=0.05$. Under (\ref{eq:spend}), each evaluation and calibration allocation is $3\alpha/(\pi^2R)$. For each metric, the controller collects independent confirmatory and calibration samples, computes $L^E_{j,r}$ from (\ref{eq:eval-lcb}), computes $U^b_{j,r}$ from (\ref{eq:bias-ucb}), and sets $L^R_{j,r}=L^E_{j,r}-U^b_{j,r}$. The candidate passes only if the primary lower bound exceeds its prespecified meaningful-improvement margin and both protected lower bounds exceed their respective non-inferiority thresholds. This example is procedural: the margins must be selected from application requirements, not tuned after observing results.

A positive observed operational-evaluator mean is insufficient. If the evaluator discrepancy is large or poorly calibrated, the candidate may remain inconclusive even when the operational score is high. Likewise, failure of one protected test blocks acceptance even if the primary objective improves substantially. Governance checks are evaluated separately and cannot be offset by a numerical score.

\end{document}